\documentclass{article}

\usepackage[preprint]{neurips_2026}

\usepackage[utf8]{inputenc} %
\usepackage[T1]{fontenc}    %
\usepackage{hyperref}       %
\usepackage{url}            %
\usepackage{booktabs}       %
\usepackage{amsfonts}       %
\usepackage{nicefrac}       %
\usepackage{microtype}      %
\usepackage[svgnames]{xcolor}         %

\usepackage{graphicx}
\usepackage{multirow}
\usepackage{amsmath}
\usepackage{algorithm}
\usepackage{algpseudocode}
\usepackage{ltl}

\usepackage{adjustbox}

\usepackage{listings}
\lstdefinestyle{framedstyle}{
    basicstyle=\ttfamily\small,
    numbers=left,
    numberstyle=\small,
    numbersep=10pt,
    frame=lines,              %
    framesep=10pt,
    xleftmargin=2em,
    framexleftmargin=2em,
    breaklines=true,
    breakindent=0pt,
    title=\lstname,
    captionpos=t,
}

\lstdefinestyle{verilogstyle}{
    language=Verilog,
    basicstyle=\ttfamily\footnotesize,
    keywordstyle=\color{blue!70}\bfseries,
    commentstyle=\color{green!50!black}\itshape,
    stringstyle=\color{red!70!black},
    numberstyle=\tiny\color{gray},
    numbers=left,
    numbersep=8pt,
    stepnumber=1,
    showstringspaces=false,
    breaklines=true,
    frame=single,
    rulecolor=\color{black!30},
    tabsize=4,
    captionpos=b,
    morekeywords={posedge,negedge,always,assign,wire,reg,module,endmodule,
                  input,output,inout,begin,end,if,else}
}

\lstdefinelanguage{TLSF}{
    morekeywords={GLOBAL,MAIN,PARAMETERS,INPUTS,OUTPUTS,ASSERT,
                  ASSUME,GUARANTEE,INITIALLY, DEFINITIONS,INFO,TITLE,DESCRIPTION,
                  SEMANTICS,TARGET,Mealy,Moore},
    morekeywords=[2]{G,F,X,U,W,R}, %
    sensitive=true,
    morecomment=[l]{//},
    morecomment=[s]{/*}{*/},
    morestring=[b]",
    alsoletter={\&,|},
}

\newsavebox\solbox
\makeatletter
{
\@parboxrestore%
\begin{lrbox}{\solbox}%
\begin{minipage}{\dimexpr\linewidth-2\fboxsep-2\fboxrule}%
}
{\end{minipage}\end{lrbox}
\framebox{\usebox\solbox}
}
\makeatother

\lstdefinestyle{tlsfstyle}{
    language=TLSF,
    basicstyle=\ttfamily\footnotesize,
    keywordstyle=\color{blue!70}\bfseries,
    keywordstyle=[2]\color{orange}\bfseries,
    commentstyle=\color{green!50!black}\itshape,
    stringstyle=\color{red!70!black},
    numberstyle=\tiny\color{gray},
    numbers=left,
    numbersep=8pt,
    showstringspaces=false,
    breaklines=true,
    frame=single,
    rulecolor=\color{black!30},
    tabsize=2,
    literate={->}{{$\rightarrow$}}2
             {<=}{{$\leq$}}2
             {!}{{$\neg$}}1
}

\usepackage{xspace}
\usepackage{wrapfig}
\usetikzlibrary{positioning, arrows.meta, calc}

\usepackage{pgfplots}
\usepackage{pgfplotstable}
\pgfplotsset{compat=1.18}

\newcommand{\MAXPARAM}{MAX\_PARAM\xspace}
\newcommand{\NL}{NATURAL\xspace}

\title{Natural Synthesis: Outperforming Reactive Synthesis Tools with Large Reasoning Models}

\author{%
  Frederik Schmitt$^1$ \quad Matthias Cosler$^1$ \quad Niklas Metzger$^1$ \quad Julian Siber$^1$ \quad \\
  \bf Vladimir Krsmanović$^1$ \quad Mohamed Ghanem$^1$ \quad Bernd Finkbeiner$^{1,2}$ \quad \\
  $^1$CISPA Helmholtz Center for Information Security \quad $^2$Technical University of Munich\\
  \texttt{firstname.lastname@cispa.de}
}

\begin{document}

\maketitle

\begin{abstract}
Reactive synthesis, the problem of automatically constructing a hardware circuit from a logical specification, is a long-standing challenge in formal verification. It is elusive for two reasons: It is algorithmically hard, and writing formal specifications by hand is notoriously difficult.
In this paper, we tackle both sides of the problem. 
For the algorithmic side, we present a neuro-symbolic approach to reactive synthesis that couples large reasoning models with model checkers to iteratively repair a synthesized Verilog implementation via sound symbolic feedback.
Our approach solves more benchmarks than the best dedicated tools in the annual synthesis competition and extends to constructing parameterized systems, a problem known to be undecidable.
On the specification side, we introduce an autoformalization step that shifts the specification task from temporal logic to natural language by introducing a hand-authored dataset of natural-language specifications for evaluation.
We demonstrate performance comparable to that of starting from formal specifications, establishing \textit{natural synthesis} as a viable end-to-end workflow.
\end{abstract}

\section{Introduction}
\label{sec:intro}

Correctness assurance accounts for a significant portion of the hardware design process~\citep{Foster2022}, yet it is indispensable, as bugs in hardware cannot be patched after production. The go-to solutions in the hardware industry are based on testing~\citep{TrippelSCKRH22} and formal verification~\citep{Clarke2018}. Testing checks whether sampled executions from a given system design conform to the system specification, while formal verification mathematically proves that all executions are conformant. Both methods require manual drafting of the design and manual bug-fixing if the draft fails the correctness checks. In contrast, these labor-intensive steps are cut out completely by \emph{reactive synthesis}, an approach that directly generates a correct design from a given formal specification. Designers can then focus on specifying \emph{what} a system should do, rather than \emph{how} the system should do it.

Significant research has been invested into reactive synthesis algorithms and tools, but despite its alluring promise of increased design quality and decreased development costs, it has seen only limited adoption by industry~\citep{BloemGJPPW07}. To a large degree, this can be attributed to its reputation as a theoretical problem of high computational complexity (it is 2-EXPTIME-complete for linear-time temporal logic as shown by~\citet{DBLP:conf/popl/PnueliR89}) and the absence of dedicated tools that scale to industrial-sized systems. Recently, however, inquiries into neural models for reactive synthesis have shown that this algorithmic cost can be avoided in many cases~\citep{schmitt2021neural,cosler2023iterative}. These neuro-symbolic approaches build on the idea of a guess-and-check loop: Neural models generate a candidate design, which is subsequently verified by a symbolic model-checking procedure, i.e., an exhaustive verification algorithm that returns a counterexample if the design does not satisfy the specification.

\begin{figure}[t]
    \centering
    \begin{minipage}{0.48\textwidth}
\begin{lstlisting}[style=tlsfstyle, showlines=true]
GLOBAL {
  PARAMETERS { n = 27; }
}
MAIN {
  INPUTS { finished[n]; } 
  OUTPUTS { allFinished; }
  INITIALLY {
    &&[0<=i<n](!allFinished W 
               finished[i]); 
    }
  ASSERT {
    G !allFinished
      -> ||[0<=i<n] G !finished[i];
    &&[0<=i<n] (allFinished -> 
      X (!allFinished W 
         finished[i]));}
}

\end{lstlisting}

    \end{minipage}
    \hfill
    \begin{minipage}{0.46\textwidth}
\begin{lstlisting}[style=verilogstyle]
module solution (
    input clk,
    input [26:0] finished,
    output allFinished );
    reg [26:0] seen = 27'b0;
    wire [26:0] next;
    wire complete;
    assign next = seen | finished;
    assign complete = &next;
    assign allFinished = complete;
    always @(posedge clk) begin
        if (complete) begin
            seen <= 27'b0;
        end else begin
            seen <= next;
        end
    end
endmodule
\end{lstlisting}
    \end{minipage}\\
    \begin{minipage}{0.97\textwidth}
    \centering
\begin{lstlisting}[
    basicstyle=\small,
    frame=single,
    rulecolor=\color{black!30},escapeinside={(*@}{@*)},
    framesep=10pt,]
(*@\parbox[t]{\linewidth}{\linespread{1}\selectfont NL Specification: The system receives input signals of n clients given as a bitvector \texttt{finished} of length n and provides a single-bit output called \texttt{allFinished}. This output must only be enabled after all clients have set their respective input in \texttt{finished} since the last time \texttt{allFinished} has been enabled. Moreover, in all cycles, the output must be enabled in the current or a future cycle if all clients set their respective input in the current or in a future cycle. The parameter n must be set to 27.} @*)
\end{lstlisting}
    \end{minipage}
    \caption{We show an example temporal specification in TLSF format for a completion detector in the upper left. A synthesized Verilog implementation for the specification is provided in the upper right. At the bottom, we show a natural language specification that can be used as input to our approach.}
    \label{fig:reactive-synthesis}
\end{figure}

In this paper, we study the capabilities of Large Reasoning Models (LRMs) for the reactive synthesis problem. We show that, already out-of-the-box, modern LRMs partially outperform dedicated tools for reactive synthesis on the problems of the annual competition on reactive synthesis SYNTCOMP~\citep{DBLP:journals/sttt/JacobsPABCCDDDFFKKLMMPR24}. We propose a method that significantly widens this gap by exploiting the feedback generated in a guess-and-check loop: By using the counterexample generated by the model checker as an input in the subsequent reasoning step, we improve the performance of LRMs and solve 92\% of the SYNTCOMP benchmarks, compared to 82\% for the best symbolic tools.

As input, our approach takes synthesis problems in the Temporal Logic Synthesis Format (TLSF), also used in SYNTCOMP. Such a problem may look like the example depicted in the upper left of Figure~\ref{fig:reactive-synthesis}. This TLSF specification describes a completion detector, which receives signals of 27 clients via the input bit-vector \texttt{finished} and produces the output \texttt{allFinished} when all clients have sent an input at least once since the last time the output was set. The TLSF specification explicitly specifies inputs and outputs, as well as an initial condition (via {\color{blue!70}\bfseries\texttt{INITIALLY}}) and an invariant (via {\color{blue!70}\bfseries\texttt{ASSERT}}). The specification is parameterized, i.e., the number of clients can be easily adjusted by changing the parameter $n$, and classic synthesis approaches then generate a design for a fixed parameter. The designs produced by our synthesis approach are given in the high-level hardware description language Verilog as shown in the upper right of Figure~\ref{fig:reactive-synthesis}. Verilog~\citep{verilog} is an industry standard for the development of hardware designs and facilitates translation of a design to physical hardware. There is also extensive tool support that allows us to model-check the generated Verilog code against the original TLSF specification to obtain feedback for the LRM in case of an incorrect generation.

The strong performance of the LRM approach on classic reactive synthesis problems motivates us to tackle problems that are traditionally considered out of reach for algorithmic approaches, but would be of high practical significance if solved.
First, we consider the parameterized synthesis~\citep{DBLP:journals/corr/JacobsB14} problem, which asks for a system design that is independent of the exact number of processes and can be instantiated with the parameter $n$ after synthesis (cf. the parameter $n$ in Figure~\ref{fig:reactive-synthesis}). Parameterized synthesis is known to be undecidable, but it would help in speeding up the development process by generating hardware designs that are independent of changing physical constraints. We show that our approach can largely generalize the reasoning from fixed parameter values to parameterized synthesis by evaluating on a subset of the SYNTCOMP data with the task of generating parameterized implementations. 
Second, we consider the problem of generating Verilog code from specifications given in natural language, as shown at the bottom of Figure~\ref{fig:reactive-synthesis}. Such synthesis from natural language may help in the challenging and time-consuming step of formalizing the system specification in a logic like TLSF. We show that our approach based on LRMs is remarkably successful at synthesizing Verilog code straight from natural language. Moreover, if desired, the approach can also first generate a formal specification before moving on to reactive synthesis. This allows designers to check the autoformalization process or verify the generated system against this specification.

A brief summary of our contributions and outline of the paper:

\begin{enumerate}
        \item We combine Large Reasoning Models with feedback from formal verification tools to generate provably correct hardware circuits from temporal-logic specifications. This counterexample-guided LRM approach (CEX-LRM) clearly outperforms state-of-the-art symbolic synthesis on standardized competition benchmarks (Section~\ref{sec:reactive-synthesis}).
    \item We apply CEX-LRM to parameterized synthesis, an undecidable problem where a system design must be able to handle an arbitrary number of clients. Despite the increased complexity, our approach performs almost equally well as in the non-parameterized case (Section~\ref{sec:parameterized-synthesis}).
    \item We extend CEX-LRM to handle natural-language specifications and evaluate it on a new, handcrafted dataset. Our \emph{natural synthesis} approach can first return an autoformalized specification for inspection, but also directly synthesize a design starting from natural language. We show that even when handicapped in this way, natural synthesis can solve challenging benchmarks out of scope for purely symbolic methods (Section~\ref{sec:natural-synthesis}).

\end{enumerate}

\section{Background}
\label{sec:background}

Reactive synthesis~\citep{Church/57/Applications} is a core problem in theoretical computer science, with foundational solutions dating back to the 1960s~\citep{Buchi1969SolvingSC}.
A common formulation is synthesis from temporal specifications, e.g., linear-time temporal logic (LTL)~\citep{LTL}, where the objective is to construct a finite-state system that satisfies a given LTL formula $\varphi$. Due to this sound-by-construction approach, reactive synthesis has the potential to revolutionize hardware design, rendering manual implementation superfluous. 
LTL augments propositional logic (operators $\neg, \wedge,\vee,\rightarrow$) with temporal operators such as $X$ (next), $U$ (until), $G$ (always), and $\LTLeventually$ (eventually), enabling the specification of behaviors for systems that continuously interact with an environment.
For more concise specifications and to express parameterized synthesis problems, LTL is usually substituted by the Temporal Logic Synthesis Format (TLSF) in practice. TLSF allows convenient constructs such as parameter declarations (cf.\ line 2 in the top left part of Figure~\ref{fig:reactive-synthesis}), succinct operators defined over ranges (cf.\ line 13 in the top left part of Figure~\ref{fig:reactive-synthesis}), and function definitions. For a fixed parameter, a TLSF specification can be compiled down to explicit LTL. 

Implementations are usually modeled as sequential circuits $C$ that translate infinite streams of inputs into infinite streams of outputs.
A circuit satisfies a formula, denoted by $C \vDash \varphi$, if all executions $\sigma$ of the circuit satisfy the formula $\varphi$.
If $C \nvDash \varphi$, then there also exists a counterexample execution $\sigma$ of $C$ that violates $\varphi$.
We use Verilog~\citep{verilog} for synthesizing hardware implementations, a hardware description language that allows us to stay at the register-transfer level rather than the gate level. 
Verilog can directly be compiled into standard circuit representations~\citep{Yosys} and model-checked against temporal properties~\citep{DBLP:conf/cav/CavadaCDGMMMRT14}.
Techniques for reactive synthesis for circuits are typically split into game-based approaches~\citep{DBLP:reference/mc/BloemCJ18} and bounded synthesis methods~\citep{DBLP:conf/cav/FaymonvilleFT17}, and all must cope with the problem's 2-EXPTIME-completeness~\citep{DBLP:conf/popl/PnueliR89}.
State-of-the-art tools such as Strix~\citep{DBLP:conf/cav/MeyerSL18} and ltlsynt~\citep{DBLP:journals/fmsd/RenkinSDP22} rely on traditional automata and game-solving algorithms, whereas recent approaches started utilizing neural methods, either as a part of heuristic search methods in SemML~\citep{DBLP:conf/tacas/KretinskyMPZ25} or for end-to-end generation~\citep{schmitt2021neural,cosler2024portfolio,cosler2023iterative}.

\section{Counterexample-Guided Synthesis with Large Reasoning Models}
\label{sec:reactive-synthesis}

In this section, we consider the classical reactive synthesis problem: given a formal logic specification $\varphi$, find a system that either satisfies the specification or prove that no such system exists.
We adopt the evaluation setting of SYNTCOMP, which represents the cumulative result of decades of research on this problem and therefore provides a strong baseline for our experiments.
Our approach implements two fundamental changes to synthesis algorithms: We use Large Reasoning Models to synthesize implementations in \emph{hardware description language} and introduce a feedback loop via sound model-checking algorithms.

\subsection{Dataset}
\label{sec:synthesis-dataset}

We collected all benchmarks from the LTL synthesis track of the reactive synthesis competition 2025 (SYNTCOMP 2025).\footnote{\href{https://github.com/SYNTCOMP/benchmarks}{https://github.com/SYNTCOMP/benchmarks}}
The total of 1586 specifications are expressed in the Temporal Logic Synthesis Format (TLSF)~\citep{DBLP:journals/corr/abs-2303-03839}.
We strip the specification of all comments, descriptions, and their names to avoid any hints on how to solve the specification or on its realizability.
Of the 1586 specifications, 458 are known to be realizable, and 419 are known to be unrealizable based on metadata provided by the competition. 
For the remaining specifications, the status of realizability remains unknown.
In the competition, tools are given 60 minutes and 160 GB of memory to solve a single instance.
The 2025 competition was won by \texttt{ltlsynt}~\citep{DBLP:journals/fmsd/RenkinSDP22} with 1297 instances solved, closely followed by \texttt{SemML}~\citep{DBLP:conf/tacas/KretinskyMPZ25} with 1295 instances solved.

\subsection{Method}

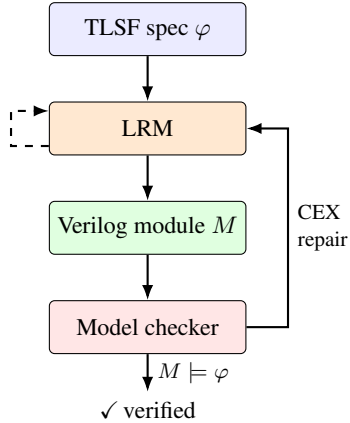
\begin{wrapfigure}{r}{0.34\linewidth}
\centering
\begin{tikzpicture}[
  font=\small,
  node distance=6mm,
  box/.style={draw, rounded corners=2pt, align=center,
              minimum width=2.6cm, minimum height=7mm, inner sep=2pt},
  arr/.style={-{Latex[length=2mm]}, thick}
]
  \node[box, fill=blue!8]                   (spec) {TLSF spec $\varphi$};
  \node[box, fill=orange!18, below=of spec] (lrm)  {LRM};
  \node[box, fill=green!12,  below=of lrm]  (ver)  {Verilog module $M$};
  \node[box, fill=red!10,    below=of ver]  (mc)   {Model checker};
  \node[below=5mm of mc, align=center]      (out)  {\checkmark\,verified};

  \draw[arr] (spec) -- (lrm);
  \draw[arr] (lrm)  -- (ver);
  \draw[arr] (ver) -- (mc);
  \draw[arr] (mc) -- node[right, font=\footnotesize] {$M \models \varphi$} (out);

  \draw[arr, dashed]
        ($(lrm.south west)!0.35!(lrm.west)$) -- ++(-0.5,0)
        |- ($(lrm.north west)!0.35!(lrm.west)$);

  \draw[arr] (mc.east) -- ++(0.55,0)
             |- node[pos=0.25, right, align=left] {\footnotesize CEX\\\footnotesize repair}
             (lrm.east);
\end{tikzpicture}
\caption{LRM-based synthesis loop: the model is prompted with the TLSF spec and emits a Verilog module, which is model-checked against the same spec; counterexamples are returned for repair.}
\label{fig:method}
\end{wrapfigure}

We begin with prompting a reasoning model to emit a Verilog module for a given specification in TLSF format.
While most reactive synthesis approaches encode the problem to an infinite game, e.g., a parity game~\citep{DBLP:journals/tcs/Zielonka98}, we intentionally stay at a high abstraction level for both specification and implementation.
Rather than translating the TLSF specification into a pure LTL formula or a Büchi automaton, we feed it directly into the LRM, preserving its programmatic structure.
Similarly, on the implementation side, we do not construct a gate-level implementation but stay on the register-transfer level and instruct the LRM to generate a Verilog module.
The succinct representations are a key aspect of our method, allowing the LRM to use more of the reasoning context window to iterate and refine solutions rather than simply representing the specification and implementation.
Note that Verilog is a suitable target language, as it can be model-checked against LTL with Yosys~\citep{Yosys}.\footnote{This only holds for a fragment of Verilog. However, this fragment is sufficient for LTL synthesis.}
We handle both realizable and unrealizable specifications by instructing the model to either generate a Verilog module satisfying the specification or to produce a Verilog module serving as an environment strategy that proves the specification is unrealizable.
In the prompt, we further add constraints on the Verilog code, the clock signal, and the overall format to simplify module extraction and support our model-checking workflow, as detailed below.
We refer to Appendix~\ref{app:synthesis-prompts} for the full prompt.

The Verilog module generated by the LRM is not correct by construction. 
However, we can automatically verify the implementation against the input specification by adding a model-checking step after inference.
The model checker then either proves correctness or constructs a counterexample trace that is an execution of the Verilog module but violates the specification.
In the latter case, we provide the counterexample to the LRM as feedback together with an instruction to repair the module.
This loop is grounded in sound symbolic feedback and adds to the LRM's unsound reasoning capabilities.
In our experiments, we repeat this process using the model-checking pipeline detailed below.

To automatically verify a Verilog module against a TLSF specification, we rely on the nuXmv model checker~\citep{DBLP:conf/cav/CavadaCDGMMMRT14} and invoke the IC3 routine~\citep{DBLP:conf/vmcai/Bradley11}.
Additionally, we have implemented a decomposition of the specification to make the problem manageable and to provide fine-grained feedback on violated sub-specifications to the LRM (see Appendix~\ref{app:decomposition} for details).
We prepare problems by translating the Verilog module into an AIGER representation with Yosys~\citep{Yosys} and then translating the AIGER representation into the model checker's input format with aigtosmv~\citep{biere2007aiger}.
The full translation script for Yosys and the model checking script for nuXmv are given in Appendix~\ref{app:model-checking-details}.
We run the model checker on an Intel Emerald Rapids machine with a 32 GB memory limit and a timeout of 600 seconds.

\subsection{Results}

We evaluated our method on the synthesis competition benchmarks with the LRMs Gemini 3.1 Pro~\citep{gemini-3-1-pro} and GPT-5.5~\citep{gpt-5-5}.
We configured both models with their highest possible reasoning setting (\texttt{HIGH} for Gemini 3.1 Pro and \texttt{XHIGH} for GPT-5.5).
As baselines, we compare with the winner of the 2025 synthesis competition \texttt{ltlsynt}~\citep{DBLP:journals/fmsd/RenkinSDP22} and the runner-up \texttt{SemML}~\citep{DBLP:conf/tacas/KretinskyMPZ25}.
Additionally, we compare against the results of \citet{DBLP:journals/corr/abs-2603-20264}; however, we note the substantial differences with respect to realizability, hardware representations, and counterexample-guidance.

We present the overall results in Table~\ref{table:reactive-synthesis}.
Both LRMs clearly outperform state-of-the-art reactive synthesis tools, with GPT-5.5 performing best and solving 170 more instances than the winner of the reactive synthesis competition.
The results show that the models benefit from sound feedback from a verification tool.
The effect is most pronounced for the Gemini model, with 162 additional instances solved after including the counterexamples provided by the model checker.
In contrast with \citet{DBLP:journals/corr/abs-2603-20264}, we arrive at a different conclusion on how LRMs can support solving reactive synthesis problems as a result of including unrealizable specifications, generating Verilog, providing counterexamples, and using higher reasoning budgets.
With respect to realizability in particular, further inspection of the results reveals that our method handles realizable and unrealizable specifications equally well.
For example, the 1392 specifications solved by GPT-5.5 split into 640 realizable and 752 unrealizable specifications.

\begin{figure}[h]
  \centering
  \hspace{-1cm}
  \begin{tikzpicture}
    \begin{axis}[
      width=0.85\linewidth,
      height=4.5cm,
      xmode=log,
      xlabel={Average use of reasoning tokens (log scale)},
      ylabel={Solved instances},
      ymin=500, ymax=1586,
      grid=major,
      grid style={dashed, gray!30},
      legend pos=south east,
      legend cell align=left,
      nodes near coords style={font=\scriptsize},
    ]
      \addplot[
        only marks, mark=*, color=blue,
        nodes near coords,
        point meta=explicit symbolic,
        every node near coord/.append style={anchor=south, yshift=2pt, color=DarkBlue!70!black},
      ] table[meta=label] {
        x     y     label
        1082  630   LOW
        3490  903   MEDIUM
        14841 1193  HIGH
      };
      \addlegendentry{\texttt{Gemini 3.1 Pro}}
      \addplot[DarkBlue, thick, no marks, forget plot] table[
        y={create col/linear regression={y=y}}
      ] {
        x     y
        1082  630
        3490  903
        14841 1193
      };
    
      \addplot[
        only marks, mark=square*, color=red,
        nodes near coords,
        point meta=explicit symbolic,
        every node near coord/.append style={anchor=north, yshift=-3pt, color=red!70!black},
      ] table[meta=label] {
        x     y     label
        1025  911   LOW
        3450  1287  MEDIUM
        5343  1378  HIGH
        6892  1392  XHIGH
      };
      \addlegendentry{\texttt{GPT-5.5}}
      \addplot[red, thick, no marks, forget plot] table[
        y={create col/linear regression={y=y}}
      ] {
        x     y
        1025  911
        3450  1287
        5343  1378
        6892  1392
      };
    \end{axis}
  \end{tikzpicture}
  \caption{Solved instances on SYNTCOMP vs.\ average reasoning tokens, with per-model linear fit.}
  \label{fig:reasoning-ablation}
\end{figure}
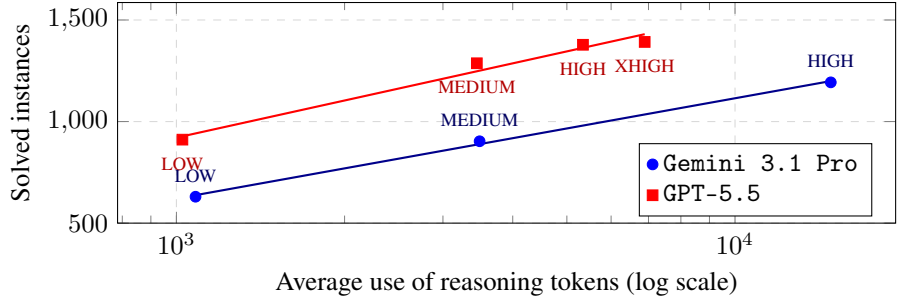

In an ablation study, we evaluated how much each LRM's reasoning token budget contributed to finding correct solutions to the synthesis problems.
We compared the different configurations for controlling the number of reasoning tokens that each model provides.
In Figure~\ref{fig:reasoning-ablation}, we plot for each configuration the number of solved instances and the corresponding average number of reasoning tokens spent.
For a full overview of the numbers, we refer to Table~\ref{table:full-reasoning} in Appendix~\ref{app:reasoning-ablations}.
We observe a clear trend: increased reasoning token budgets are associated with a higher number of solved instances.
More specifically, the trend is log-linear for each model, suggesting diminishing returns for very high reasoning budgets.
GPT-5.5 additionally supports a configuration with no reasoning tokens at all.
In that case, the number of solved instances drops sharply to 123.
Based on these results, we can attribute the ability to find correct solutions to synthesis problems largely to advances in reasoning models and to the number of reasoning tokens that are spent.

\begin{table}[bp]
  \caption{Reactive synthesis results for competition benchmarks: comparing LRMs with algorithms \texttt{ltlsynt} and \texttt{SemML}. LRMs were run with their highest reasoning configurations.}
  \label{table:reactive-synthesis}
  \centering
  \begin{tabular}{llc r@{\,/\,}l}
    \toprule
    Type & Name & CEX Iterations & \multicolumn{2}{c}{Solved} \\
    \midrule
    \multirow{2}{*}{Algorithm} & \texttt{SemML}~\citep{DBLP:conf/tacas/KretinskyMPZ25} & & 1295 & 1586 \\
    & \texttt{ltlsynt}~\citep{DBLP:journals/fmsd/RenkinSDP22} & & 1297 & 1586 \\
    \cmidrule{1-5}
    \multirow{7}{*}{LRM} & \texttt{GPT-5}~\citep{DBLP:journals/corr/abs-2603-20264}  & &  229 & 1586 \\
    \cmidrule{2-5}
    & \multirow{3}{*}{\texttt{Gemini 3.1 Pro}} & 0 & 1193 & 1586 \\ %
    & & 1 & 1311 & 1586 \\
    & & 2  & 1355 & 1586 \\
    \cmidrule{2-5}
    & \multirow{3}{*}{\texttt{GPT-5.5}} & 0 & 1392 & 1586 \\ %
    & & 1  & 1449 & 1586 \\ %
    & & 2 & $\mathbf{1467}$ & 1586 \\ %
    \bottomrule
  \end{tabular}
\end{table}

\section{Reactive Synthesis Beyond Decidability}
\label{sec:parameterized-synthesis}
The previous experiments showed that the natural synthesis framework outperforms existing symbolic tools for LTL synthesis.
In this section, we show that Large Reasoning Models can lift the problem to the next level:
we synthesize Verilog modules for LTL formulas with variable numbers of input and output variables, as well as variable numbers of operator sequences.
This elevates the task to the undecidable problem of \emph{parameterized synthesis}~\citep{DBLP:journals/corr/JacobsB14}, yet the LRM can still synthesize correct parameterized Verilog implementations.

Parameterized synthesis generalizes the classical reactive synthesis problem: given a specification parameterized in the number of processes, find an implementation template whose instantiations satisfy the specification regardless of the number of processes or prove that no such template exists.
Importantly, the generalization makes the synthesis problem computationally much more challenging.
In fact, for linear-time temporal logic, the generalization turns the problem into an undecidable problem~\citep{DBLP:journals/corr/JacobsB14}.
Yet, finding reusable and configurable, parameterized implementations is of great interest to practitioners.
Consider, for example, the detector in Figure~\ref{fig:reactive-synthesis}.
Ideally, we want the implementation to be independent of the number of clients
and to work correctly for an arbitrary number.
Verilog directly supports such generalizations, for example through parameter declarations in the module's header list.
In Figure~\ref{fig:parameterized-detector} in Appendix~\ref{app:parameterized-synthesis} we show the parameterized version of the detector.
In the following, we present how our approach can be directly extended to find such parameterized implementations.

\subsection{Dataset}
\label{sec:parameterized-dataset}

The synthesis competition benchmarks introduced in Section~\ref{sec:synthesis-dataset} already contain many specifications that are equivalent up to a parameter value.
The detector benchmark shown in Figure~\ref{fig:reactive-synthesis} is an example of such a specification.
In addition to parameter value $n=27$, it is contained in the competition benchmarks for nine smaller parameter values.
To evaluate our parameterized synthesis approach, we identify such benchmarks and derive a dataset of 57 general parameterized specifications.
Similar to the competition benchmarks themselves we post-process them by removing all comments, descriptions, and names to avoid any hints on how to solve the specification.

\subsection{Method}
Our method generally follows the method described for reactive synthesis in Section~\ref{sec:reactive-synthesis}.
We modify the LRM instruction to generate a parameterized Verilog implementation with a parameter declaration in the Verilog module's header list.
Importantly, we can no longer automatically verify the parameterized implementation since the parameterized LTL verification is undecidable similar to the synthesis problem~\citep{DBLP:series/synthesis/2015Bloem}.
We therefore resort to testing different parameter values and verify for each value the instantiated Verilog module against the instantiated specification similar to the reactive synthesis case.
We test up to the largest parameter value found in the synthesis competition for the respective specification class.
In case we find a violation for one parameter value, we still obtain a sound counterexample that we provide to the LRM as feedback.
However, we can no longer guarantee correctness in the positive case.

\subsection{Results}

As a consequence of the undecidability of the problem, tools and algorithms for solving parameterized synthesis problems are scarce.
To the best of our knowledge, no tool exists that is mature enough to serve as a baseline for the problem set introduced above.
We therefore only evaluate our approach without a baseline comparison in this section.
We use Gemini 3.1 Pro~\citep{gemini-3-1-pro} and GPT-5.5~\citep{gpt-5-5} configured with their highest possible reasoning settings.
The results for both models are presented in Table~\ref{table:parameterized}.
With an average of 35 instances solved, both models perform on par.
However, the Gemini model benefits more from additional verification feedback, solving up to 38 instances.
Further, the task of finding a parameterized implementation seems only moderately harder for the LRMs than finding an implementation for a specific parameter value.
For comparison, the Gemini model solves 38 instances, and the GPT model 41, on average, when instantiating the problems with the largest parameter value found in the synthesis competition (See Appendix~\ref{app:max-param}).

\begin{table}[t]
  \caption{Parameterized synthesis results for \texttt{Gemini 3.1 Pro} and \texttt{GPT-5.5}. We report the average over three runs with standard deviation.}
  \label{table:parameterized}
  \centering
  \begin{tabular}{lr@{\,/\,}l}
    \toprule
    Name & \multicolumn{2}{c}{Solved} \\
    \midrule
    \texttt{Gemini 3.1 Pro} & $35.0\pm1.0$ & 57 \\
    + \texttt{CE-Guidance} & $\mathbf{38.3}\pm1.1$ & 57 \\
    \cmidrule{1-3}
    \texttt{GPT-5.5} & $35.7\pm2.5$ & 57 \\
    + \texttt{CE-Guidance} & $35.7\pm2.5$ & 57 \\
    \bottomrule
  \end{tabular}
\end{table}

\section{Reactive Synthesis Beyond Temporal Logic}
\label{sec:natural-synthesis}

\begin{wrapfigure}{r}{0.4\linewidth}
\vspace{-1em}
\centering
\begin{tikzpicture}[
  font=\small,
  node distance=4mm,
  box/.style={draw, rounded corners=2pt, align=center,
              minimum width=2.6cm, minimum height=7mm, inner sep=2pt},
  sbox/.style={box, minimum width=2.3cm, minimum height=7mm},
  arr/.style={-{Latex[length=2mm]}, thick}
]
  \node[box, fill=blue!8]                    (nl)   {NL description};
  \node[box, fill=orange!18, below=of nl]    (auto) {LRM};
  \node[box, fill=blue!8,   below=of auto]  (tlsf) {TLSF spec $\hat\varphi$};
  \node[box, fill=orange!18, below=of tlsf]  (syn)  {LRM};
  \node[box, fill=green!12,  below=of syn]   (ver)  {Verilog module $M$};

  \node[box, fill=red!10, below=6mm of ver] (mc1) {Model checker};

  \node[below=5mm of mc1, align=center]      (out)  {\checkmark\,verified};

  \draw[arr] (mc1) -- node[right, font=\footnotesize] {$M \models \hat\varphi$} (out);

  \draw[arr] (nl)   -- (auto);
  \draw[arr] (auto) -- (tlsf);
  \draw[arr] (tlsf) -- (syn);
  \draw[arr] (syn)  -- (ver);
  \draw[arr] (ver.south) -- ++(0,-0.2) -| (mc1.north);

  \draw[arr, dashed]
        (nl.south east) to[out=-60, in=60]
        node[right, font=\footnotesize, pos=0.5] {direct}
        (syn.north east);

  \draw[arr] ([yshift=1mm]tlsf.west) -- ++(-0.3,0)
             |- node[pos=0.25, left, align=right, font=\footnotesize] {syntax\\repair}
             (auto.west);
\end{tikzpicture}
\caption{The autoformalization route (solid) translates NL to TLSF $\hat\varphi$, then synthesizes Verilog. The direct route (dashed) skips autoformalization.
The module is verified against the autoformalized specification.
}
\label{fig:natural-synthesis}
\vspace{-2.25em}
\end{wrapfigure}
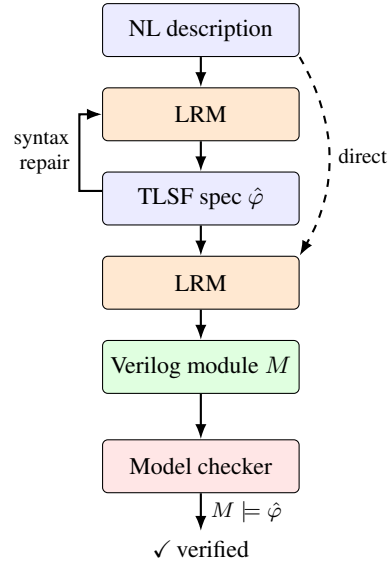
Writing formal specifications in temporal logic is notoriously difficult and requires expertise in formal verification.
Describing a system's desired behavior in natural language is often more intuitive and accessible, but out of scope for symbolic algorithms.
With Large Reasoning Models, we can now address the entire reactive-synthesis pipeline -- from informal requirements to formal temporal-logic specifications to verified implementations -- end-to-end.
We manually authored a dataset of natural language specifications on which we perform natural synthesis with two approaches:
1) We autoformalize the natural language specification into a formal specification in TLSF format and then synthesize a Verilog module from the formal specification (as in Section~\ref{sec:reactive-synthesis}). 
2) We prompt the LRM to directly synthesize a Verilog module from the natural language specification. 
We verify against both the correct, manually crafted specification, and the autoformalization, and measure the correctness of the autoformalization step.

\subsection{Dataset}

Based on the dataset introduced in Section~\ref{sec:parameterized-dataset}, we manually author a natural language description for each of the 57 parameterized specifications, which we call \NL.
For this experiment, we built a challenging set of benchmarks by setting the parameter values to the largest values occurring in SYNTCOMP. 
The best algorithmic tools can synthesize only 19 of 57 formalized specifications, whereas the presented approach with LRMs (Section~\ref{sec:reactive-synthesis}) solves $38.3\pm0.6$/57. 
See Appendix~\ref{app:max-param} for more details.
In this section, however, we focus on the natural language specifications.
Because the formal specifications of this dataset are already very challenging, it is a valuable testbed for evaluating the performance of synthesis from natural-language specifications.

\begin{table}[bp]
  \caption{Reactive synthesis from natural language on the manually authored dataset \NL.
  We report the mean over three runs with standard deviation. All experiments are run with Gemini 3.1 Pro and reasoning level high.}
  \label{table:natural-synthesis}
  \centering
  \begin{tabular}{llc}
    \toprule
    Approach & Verified against & Solved \\
    \midrule
    \multirow{2}{*}{Via Autoformalization} & ground truth & $30.0\pm0.0$ / 57 \\ %
    & autoformalized & $33.7\pm1.7$ / 57 \\ %
    \midrule
    \multirow{2}{*}{End-to-End} & ground truth & $31.7\pm1.2$ / 57 \\ %
    & autoformalized & $30.7\pm1.9$ / 57 \\ %
    \bottomrule
  \end{tabular}
\end{table}

\subsection{Results}
We evaluate the full natural synthesis pipeline on the \NL dataset, comparing two routes: autoformalization followed by LRM-based synthesis from these autoformalized specifications, and direct end-to-end LRM-based synthesis from natural language. We further address the correctness of the autoformalization step and its impact on synthesis performance. We verify the generated Verilog modules against both the ground truth specifications and an autoformalized specification to isolate the effects of potential semantic shifts during autoformalization.
Table~\ref{table:natural-synthesis} summarizes the results. Both approaches solve roughly 30 out of 57 specifications against the ground truth, demonstrating that LRMs can produce provably correct hardware directly from informal descriptions.

\paragraph{Autoformalization.}
We autoformalize each natural language specification into TLSF by prompting the LRM with a structured template that includes the full TLSF grammar, key conventions for parameterized signals, and LTL operators.
The LRM obtains sound feedback on the syntax of the produced specification in a feedback loop.
Since autoformalization is a challenging task, we find that the LRM cannot always produce a syntactically correct TLSF specification within three attempts ($12.7\pm0.5$/57).
We then use Spot's~\citep{DBLP:conf/cav/Duret-LutzRCRAS22} \texttt{ltlfilt} to check for equivalence with the ground truth specification.
$9.3\pm0.6$/57 are proven equivalent, while $8.7\pm0.6$/57 are provably inequivalent.
Note that, because the specifications in \NL are designed to be challenging, the automata-based equivalence check limits our evaluation ($26.3\pm0.9$/57 timed out after 30 minutes, 32GB per specification).
Autoformalization accuracies need careful evaluation: syntactically incorrect specifications are often close to the intended behavior and can still yield correct results via natural synthesis.
In fact, several specifications with syntactically broken autoformalizations still led to correct Verilog modules in subsequent experiments, indicating that the underlying semantic intent is largely preserved.
Additionally, specifications that are inequivalent to the ground truth are not necessarily incorrect, since natural language descriptions can be ambiguous and underspecified, possibly leading to sound strengthening of the specification.

\paragraph{Natural Synthesis via Autoformalized Specifications.}

To complete the pipeline, we synthesize a Verilog module from the autoformalized specification with the same method as in Section~\ref{sec:reactive-synthesis}.
We find that the syntax repair loop can introduce a semantic shift, as the LRM then focuses on syntactic correctness rather than the specification's semantics.
At the same time, during synthesis, the LRM can handle (or rather ignore) syntax errors in the generated TLSF specifications and still produce correct Verilog modules.
We therefore use the first-try autoformalized specification for synthesis, even if it contains syntax errors.
We verify the generated module against the autoformalized specification. Here, the syntax-repaired specifications become advantageous, since syntactic errors in the specification would cause verification to fail outright.
To test our method and isolate the potential semantic shift, we also verify the generated module against the ground-truth specification.
The results are shown in Table~\ref{table:natural-synthesis}. We find 30 out of 57 specifications to satisfy the ground truth, meaning they are provably correct.
$33.7\pm1.7$ out of 57 specifications satisfy the autoformalized specification, with a high number of $25.7\pm0.6$ satisfying both the ground truth and the autoformalized specification, indicating that the autoformalization step is often precise enough to specify the intended behavior fully, and a semantic shift is not a dominant issue.

\paragraph{End-to-End Natural Synthesis.}

As an alternative to the autoformalization route, we prompt the LRM to directly synthesize a Verilog module from the natural language description, bypassing the intermediate TLSF specification.
We verify the generated module against an autoformalized specification from the previous experiment,
and against the ground truth specification to isolate the effect of potential semantic shifts in the autoformalization step.
The direct approach solves $31.7\pm1.2$ out of 57 specifications against the ground truth (Table~\ref{table:natural-synthesis}), slightly outperforming the autoformalization route ($30.0\pm0.0$).
In contrast to the autoformalized specification, the direct approach solves slightly fewer instances ($30.7\pm1.9$), since the Verilog module was synthesized independently of it.
These results suggest that the intermediate formalization step does not provide a significant advantage for synthesis on this dataset.
We suspect that the specific format of TLSF adds to the complexity.
Since these models reason in natural language, staying in this modality as long as possible is likely beneficial.
Nevertheless, the performance difference between the two approaches is small.
Most importantly, generating a formal specification as an intermediate step is helpful and necessary for interpretability during hardware engineering, enabling specification refinement (e.g., resolving ambiguity, correcting errors) and formal verification.

\section{Related Work}
\label{sec:related-work}

\paragraph{Deep Learning for Hardware Design.}
Deep learning methods have been applied to many aspects of the hardware design process.
For formal verification, deep neural networks have been used as proof certificates in the form of neural ranking functions~\citep{DBLP:conf/nips/GiacobbeKPT24}.
For optimization, deep reinforcement learning has been applied to abstraction levels ranging from Boolean circuit minimization~\citep{DBLP:conf/iclr/Chowdhury00KG24,DBLP:conf/nips/WangWYBLCHYL0024} to floorplanning~\citep{DBLP:journals/nature/MirhoseiniGYJSW21}.
Similarly, research on representation learning ranges from Boolean circuits~\citep{DBLP:conf/aspdac/NetoMAYG21,DBLP:conf/iclr/deepgate4} to register-transfer-level abstractions~\citep{DBLP:conf/nips/VasudevanJBSSHS21}.
For hardware description languages in particular, language-modeling techniques were explored both for the Verilog language~\citep{DBLP:journals/todaes/ThakurAPTDKG24, DBLP:journals/corr/abs-2505-24183} and for the translation from natural language to Verilog~\citep{DBLP:conf/mlcad/PearceTK20}.
The reactive synthesis problem itself has been studied as a deep learning translation from LTL to gate-level circuits~\citep{schmitt2021neural,cosler2023iterative}
and combined with symbolic solvers in a neural-symbolic portfolio solver~\citep{cosler2024portfolio}.
Closest to our work are~\citet{DBLP:journals/corr/abs-2603-20264}, who compare LRMs with limited reasoning against symbolic synthesis tools on the task of synthesizing realizable LTL specifications to gate-level circuits, not a hardware description language like Verilog that operates at the register-transfer level.

\paragraph{Autoformalization with LLMs.}
Autoformalization is the task of translating informal, natural language into a formal logic or formal mathematical statement. %
Recently, it has been extensively studied in theorem proving with progress driven by Large Language Models (LLMs)~\citep{DBLP:conf/nips/WuJLRSJS22,DBLP:conf/iclr/JiangWZL0LJLW23,DBLP:conf/icml/MurphyYSLAS24,jana2025proofbridge}.
Closely related to theorem proving, it has proven to be a crucial component of AI systems that perform well in math competitions~\citep{1062014}.
Similar to the developments in formal mathematics, LLMs have fueled research in autoformalization in formal verification.
For temporal logic specifically, the formalization of unstructured natural language into LTL formulas has been studied as an end-to-end translation task~\citep{DBLP:conf/aaai/FuggittiC23,DBLP:conf/emnlp/ChenGZF23,DBLP:conf/corl/LiuYILSTS23}, via the composition of sub-translations~\citep{DBLP:conf/cav/CoslerHMST23,DBLP:conf/fmcad/MendozaHT24}, via separation of data and control~\citep{DBLP:journals/corr/abs-2406-07400}, and constrained decoding~\citep{DBLP:conf/icml/EnglishS0E25}.
The recently introduced VerifyThisBench targets the translation of natural language into formal specifications for program verification~\citep{DBLP:journals/corr/abs-2505-19271}.

\section{Conclusion}
\label{sec:conclusion}
We introduced \emph{natural synthesis}, a neuro-symbolic approach to reactive synthesis that combines Large Reasoning Models with counterexample-guided reasoning.
Our pipeline substantially outperforms purely symbolic approaches on SYNTCOMP benchmarks, extends to parameterized (undecidable) synthesis, and supports synthesis from natural-language descriptions through autoformalization and direct circuit generation, all verified against temporal specifications.
These results highlight the potential of neuro-symbolic methods to bring reactive synthesis into real-world hardware design workflows. 
These promising observations motivate concrete next steps: creating new synthesis benchmarks based on natural-language hardware specifications to better capture practical use cases, and developing scalable, potentially neuro-symbolic, hardware verification techniques to overcome model-checking bottlenecks.

\begin{ack}
This work was partially supported by the European Union with ERC Grant HYPER (No. 101055412).
Views and opinions expressed are however those of the authors only and do not necessarily reflect those of the European Union or the European Research Council Executive Agency.
Neither the European Union nor the granting authority can be held responsible for them.
\end{ack}

\bibliographystyle{abbrvnat}
\bibliography{neurips_2026}

\clearpage
\appendix

\section{Reactive Synthesis}

\subsection{Reactive Synthesis Prompt}
\label{app:synthesis-prompts}

\begin{figure}[ht]
\centering
\begin{lstlisting}[style=framedstyle]
Your task is to translate the following temporal logic specification in Temporal Logic Synthesis Format (TLSF) into a Verilog module that satisfies the specification if realizable, or represents an environment strategy if unrealizable.

The TLSF format builds upon standard Linear Temporal Logic (LTL) and breaks the specification down into up to six sections: INITIALLY, PRESET, REQUIRE, ASSERT/INVARIANTS, ASSUME/ASSUMPTIONS, and GUARANTEE/GUARANTEES. If the semantics are "Mealy" or "Moore", the sections are interpreted as the formula f_initially \rightarrow (f_preset \land (G f_require \land f_assume \rightarrow G f_assert \land f_guarantee)). If the semantics are "Mealy,Strict" or "Moore,Strict", the sections are interpreted as the formula f_initially \rightarrow (f_preset \land (f_assert W \neg f_require) \land (G f_require \land f_assume \rightarrow f_guarantee)).

Follow these guidelines for the translation from TLSF to Verilog:

- If the specification is unrealizable, swap the roles of inputs and outputs: the module's inputs become the specification's outputs, and the module's outputs become the specification's inputs. The implementation should demonstrate an environment strategy that violates the specification for every possible system response.
- When the TLSF specification contains a PARAMETERS subsection, do not generate a parameterized Verilog module, but instead directly instantiate the parameters with their given values.
- In addition to inputs and outputs, include a single clock input named "clk" in the Verilog module and nothing else.
- Make sure the code can be processed by Yosys. For example, do not declare variables inside procedural blocks like initial or always.
- Name the module simply "solution" if the specification is realizable and "environment" if the specification is unrealizable.
- Enclose the Verilog code within triple backticks (```) and specify "verilog" right after the opening set of backticks.

Here is the TLSF specification:

{specification}

\end{lstlisting}
\caption{Reactive synthesis prompt template.}
\label{fig:synthesis-prompt}
\end{figure}
\clearpage
\subsection{\MAXPARAM Dataset}
\label{app:max-param}

\begin{table}[h]
  \caption{Synthesis results on the \MAXPARAM dataset (57 hardest parameterized SYNTCOMP benchmarks, formalized variant of the \NL dataset). LRM results report the mean over three runs with standard deviation.}
  \label{table:max-param}
  \centering
  \begin{tabular}{llc}
    \toprule
     Input & Method & Solved \\
    \midrule
     \multirow{2}{*}{Algorithmic} & \texttt{SemML}~\citep{DBLP:conf/tacas/KretinskyMPZ25} & 19 / 57 \\
     & \texttt{ltlsynt}~\citep{DBLP:journals/fmsd/RenkinSDP22} & 19 / 57 \\
    \cmidrule{1-3}
    \multirow{2}{*}{LRM} & \texttt{Gemini 3.1 Pro} & $38.3\pm0.6$ / 57 \\ %
    & \texttt{GPT-5.5} & $\mathbf{41.0}\pm2.6$ / 57 \\ %
  \end{tabular}
\end{table}

\subsection{Model Checking Details}
\label{app:model-checking-details}

\begin{algorithm}
\caption{Yosys recipe for translating Verilog to AIGER}
\label{alg:yosys-recipe}
\begin{algorithmic}[1]
\State \textbf{hierarchy} \texttt{-check -top} $module\_name$ \Comment{design hierarchy}
\State \textbf{proc} \Comment{convert RTL processes to netlist}
\State \textbf{flatten} \Comment{inline submodules}
\State \textbf{opt} \Comment{coarse-grain optimization}
\State \textbf{memory}; \textbf{opt} \Comment{map memories, then re-optimize}
\State \textbf{techmap}; \textbf{opt} \Comment{rewrite, then re-optimize}
\State \textbf{dffunmap} \Comment{decompose complex flip-flops}
\State \textbf{abc} \texttt{-g AND} \Comment{map combinational logic to AND gates}
\State \textbf{delete} \texttt{-port} $module\_name$\texttt{/clk} \Comment{strip clock port}
\end{algorithmic}
\end{algorithm}

\begin{algorithm}
\caption{nuXmv script for LTL model checking with ic3}
\label{alg:nuxmv-recipe}
\begin{algorithmic}[1]
\State \textbf{read\_model} \Comment{parse the SMV input file}
\State \textbf{flatten\_hierarchy} \Comment{inline modules into a flat namespace}
\State \textbf{encode\_variables} \Comment{assign Boolean encodings to variables}
\State \textbf{build\_boolean\_model} \Comment{construct the Boolean transition system}
\State \textbf{check\_ltlspec\_ic3} \Comment{verify LTL properties via IC3/IMC}
\State \textbf{quit} \Comment{terminate the interactive session}
\end{algorithmic}
\end{algorithm}

\clearpage
\subsection{Reasoning Level Ablations}
\label{app:reasoning-ablations}

\begin{table}[h]
  \caption{Thinking level ablation on SYNTCOMP. \textit{\# Reasoning Tokens} give the average number of tokens spent with standard deviation.}
  \label{table:full-reasoning}
  \centering
  \begin{tabular}{llll}
    \toprule
    Model & Thinking Level & Solved & \# Reasoning Tokens \\
    \midrule
    \multirow{3}{*}{\texttt{Gemini 3.1 Pro}} & LOW & $630$ / 1586 & $1082\pm686$ \\ %
    & MEDIUM & $903$ / 1586 & $3490\pm2839$\\ %
    & HIGH & $1193$ / 1586 & $14841\pm9937$ \\ %
    \cmidrule{1-4}
    \multirow{5}{*}{\texttt{GPT 5.5}} & NONE & $123$ / 1586 & $0\pm0$ \\ %
    & LOW & $911$ / 1586 & $1025\pm534$\\ %
    & MEDIUM & $1287$ / 1586 &  $3450\pm2751$\\ %
    & HIGH & $1378$ / 1586 &  $5343\pm4088$\\ %
    & XHIGH & $\mathbf{1392}$ / 1586 & $6892\pm5342$\\ %
    \bottomrule
  \end{tabular}
\end{table}

\begin{table}[h]
  \caption{Thinking level ablation on \MAXPARAM dataset. \textit{\# Reasoning Tokens} gives the average number of tokens spent. \textit{Solved} reports the mean over three runs together with standard deviation.}
  \label{table:max-param-reasoning-ablation}
  \centering
  \begin{tabular}{lll}
    \toprule
    Model & Thinking Level & Solved \\
    \midrule
    \multirow{3}{*}{\texttt{Gemini 3.1 Pro}} & LOW & $21.0\pm2.0$ / 57 \\ %
    & MEDIUM & $28.7\pm0.6$ / 57 \\ %
    & HIGH & $38.3\pm0.6$ / 57 \\ %
    \cmidrule{1-3}
    \multirow{5}{*}{\texttt{GPT 5.5}} & NONE & $6.7\pm1.5$ / 57 \\ %
    & LOW & $29.0\pm3.0$ / 57 \\ %
    & MEDIUM & $36.3\pm1.5$ / 57\\ %
    & HIGH & $38.3\pm2.1$ / 57 \\ %
    & XHIGH & $\mathbf{41.0}\pm2.6$ / 57\\ %
    \bottomrule
  \end{tabular}
\end{table}

\subsection{Decomposition of Specifications}\label{app:decomposition}
The typical format of reactive synthesis specifications with $n$ assumptions and $m$ guarantees allows for a decomposition into $m$ independent problems in the realizable case and $n$ independent problems in the unrealizable case (see Figure~\ref{fig:decomposition}).

\begin{figure}[h]
\centering
\begin{minipage}{.48\textwidth}
  \centering
  \begin{align*}
        &a_1 \land \ldots \land a_n \rightarrow g_1\\
        &\vdots\\
        &a_1 \land \ldots \land a_n \rightarrow g_m
  \end{align*}
  \vspace{0.5em}
  (a) realizable case
\end{minipage}\hfill
\begin{minipage}{.48\textwidth}
  \centering
  \begin{align*}
        &\neg(a_1 \rightarrow g_1 \land \ldots \land g_m)\\
        &\vdots\\
        &\neg(a_n \rightarrow g_1 \land \ldots \land g_m)
  \end{align*}
  \vspace{0.5em}
  (b) unrealizable case
\end{minipage}
\caption{Decomposition of the model checking problem into $m$ and $n$ subproblems, respectively.}
\label{fig:decomposition}
\end{figure}
\clearpage
\section{Parameterized Synthesis}
\label{app:parameterized-synthesis}

\begin{figure}[h]
    \centering
\begin{lstlisting}[style=verilogstyle]
module solution #(
    parameter n = 27
) (
    input clk,
    input [n-1:0] finished,
    output allFinished );
    reg [n-1:0] seen = {n{1'b0}};
    wire [n-1:0] next;
    wire complete;
    assign next = seen | finished;
    assign complete = &next;
    assign allFinished = complete;
    always @(posedge clk) begin
        if (complete) begin
            seen <= {n{1'b0}};
        end else begin
            seen <= next;
        end
    end
endmodule
\end{lstlisting}
    \caption{A parameterized version of the detector implementation shown in Figure~\ref{fig:reactive-synthesis}.}
    \label{fig:parameterized-detector}
\end{figure}

\section{Limitations}\label{app:limitations}

\paragraph{Verification bottleneck.}
Model checking the generated Verilog modules against their specifications is computationally expensive, especially for large parameter values.
While model checking is orders of magnitude easier in complexity than synthesis (PSPACE vs.\ 2EXPTIME), our approach shows that LRMs push the guess-and-check paradigm to its limit.
In our experiments, a significant fraction of instances remain unresolved due to verification timeouts, meaning our reported numbers are a lower bound on synthesis capability.

\paragraph{Undecidable parameterized verification.}
Parameterized synthesis and parameterized model checking are both undecidable in general.
Our evaluation of parameterized Verilog modules therefore relies on checking individual parameter instantiations, which does not guarantee correctness for all parameter values.
A complete verification would require a proof that the implementation is correct for every possible parameter value, which no automated procedure can provide in general.

\paragraph{Autoformalization evaluation.}
Assessing the correctness of autoformalized specifications is inherently difficult.
Equivalence checking between the autoformalized and ground-truth specifications is expensive (PSPACE-complete for LTL) and timed out for a large fraction of our dataset.
Furthermore, natural language descriptions can be ambiguous, admitting multiple distinct valid formalizations.

\paragraph{LRM dependency and reproducibility.}
Our results depend on proprietary LRM APIs whose behavior may change over time.
We do not control for model versioning beyond recording the model identifiers used (\texttt{gpt-5.5-2026-04-23} and \texttt{gemini-3.1-pro-preview}).
Model providers may discontinue models, update weights or safety filters without notice, any of which could affect performance.
Furthermore, LRM outputs are non-deterministic: even with identical prompts, different runs can produce different solutions, as reflected in the variance we report across repeated runs.

\paragraph{Time and cost.}
Running large reasoning models at high token budgets across hundreds of specifications incurs substantial computational cost.
A full evaluation of the SYNTCOMP benchmark requires approximately 11M reasoning tokens for GPT-5.5 (XHIGH) and 24M for Gemini 3.1 Pro (HIGH), with additional tokens for each repair iteration (see Table~\ref{table:full-reasoning}).
A single synchronous call on one challenging instance takes about 4 minutes for GPT-5.5 (XHIGH) and 3 minutes for Gemini 3.1 Pro (HIGH).
Both providers, however, support batch inference, heavily parallelizing the calls and reducing wall-clock time.

\end{document}